%% file: PaperForReview.tex
\documentclass[10pt,twocolumn,letterpaper]{article}

\usepackage{wacv}              

\usepackage{graphicx}
\usepackage{amsmath}
\usepackage{amssymb}
\usepackage{booktabs}
\usepackage{url,multirow,bm,bbm,breqn,rotating,color,colortbl,subcaption}
\usepackage[dvipsnames]{xcolor}
\usepackage[accsupp]{axessibility}

\usepackage[pagebackref,breaklinks,colorlinks]{hyperref}

\usepackage[capitalize]{cleveref}
\crefname{section}{Sec.}{Secs.}
\Crefname{section}{Section}{Sections}
\Crefname{table}{Table}{Tables}
\crefname{table}{Tab.}{Tabs.}

\def\wacvPaperID{2562} 
\def\confName{WACV}
\def\confYear{2024}

\def\company{Zeitview }

\begin{document}

\title{ZRG: A Dataset for Multimodal \\3D Residential Rooftop Understanding}

\author{
Isaac Corley$^1{^2}$ \quad Jonathan Lwowski$^2$ \quad Peyman Najafirad$^1$\\
$^1$University of Texas at San Antonio \quad \quad $^2$Zeitview\\
{\tt\small isaac.corley@utsa.edu, jonathan.lwowski@zeitview.com, peyman.najafirad@utsa.edu}
}

\twocolumn[{%
\renewcommand\twocolumn[1][]{#1}%
\maketitle
  \centering
  \newlength{\itemheight}{
        \centering
        \includegraphics[width=1.0\linewidth]{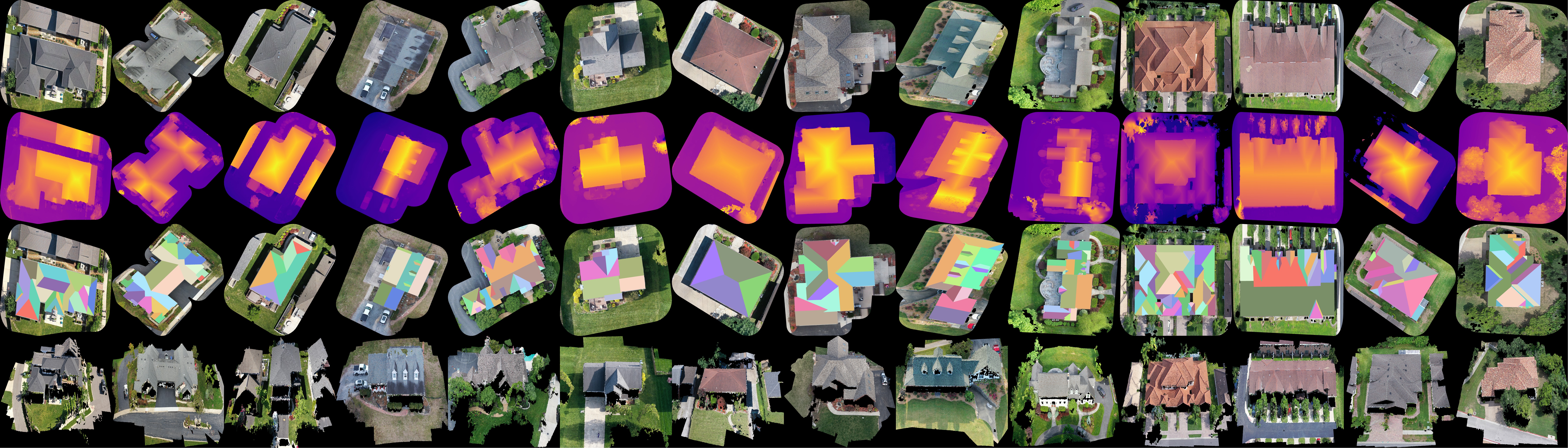}
    }
  \captionof{figure}{\textbf{Sample properties from our proposed ZRG dataset}. The dataset contains (top to bottom) high resolution RGB orthomosaics, digital surface models (DSM), 3D rooftop geometry wireframes, and 3D colored point clouds acquired from roof inspections of over 20k residential properties across the U.S.}
  \vspace{1em}
  \label{fig:sample}%
}]

\maketitle

\begin{abstract}
\input{sections/abstract}
\end{abstract}

\section{Introduction}
\label{sec:intro}
\input{sections/introduction}

\section{Background}
\label{sec:background}
\input{sections/background}

\section{Related Work}
\label{sec:related-work}
\input{sections/related_work}

\section{The ZRG Dataset}
\label{sec:dataset}
\input{sections/dataset}

\section{Experiments}
\label{sec:experiments}
\input{sections/experiments}

\section{Discussion}
\label{sec:discussion}
\input{sections/discussion}

\section{Conclusion}
\label{sec:conclusion}
\input{sections/conclusion}

{\small
\bibliographystyle{ieee_fullname}
\bibliography{egbib}
}

\end{document}

%% file: sections/abstract.tex
A crucial part of any home is the roof over our heads to protect us from the elements. In this paper we present the \company Rooftop Geometry (ZRG) dataset for residential rooftop understanding. ZRG is a large-scale residential rooftop dataset of over 20k properties collected through roof inspections from across the U.S. and contains multiple modalities including high resolution aerial orthomosaics, digital surface models (DSM), colored point clouds, and 3D roof wireframe annotations. We provide an in-depth analysis and perform several experimental baselines including roof outline extraction, monocular height estimation, and planar roof structure extraction, to illustrate a few of the numerous potential applications unlocked by this dataset.\footnote{Preprint accepted to WACV 2024}\\\\

%% file: sections/introduction.tex
The roof is a vital component of a home and serves to protect owners from various environmental elements, such as rain, snow, and wind, while also contributing to the overall aesthetics and energy efficiency of the structure. A thorough understanding of a roof's condition is critical for homeowners as it allows them to make informed decisions about repair, replacement, or maintenance and the cost thereof. However, traditional methods of roof inspection are often time-consuming, labor-intensive, and subject to human error. Rapid developments in cost-effectiveness of unmanned aerial vehicles (UAV) have created safer and more efficient alternatives to manual roof inspections which capture high-resolution images of residential roofs from various angles and perspectives. As a result, there is an increasing need to augment and/or automate the roof analysis process by combining the latest advancements in machine learning and computer vision learning with large-scale residential rooftop datasets.

\input{tables/datasets}

Given a sufficiently large residential rooftop dataset with multiple modalities, what are some of the potential applications which can be automated to further benefit society? The primary objective of this research is to answer this question by introducing a novel large-scale rooftop dataset, which we name the \company Rooftop Geometry (ZRG) dataset, for the analysis and understanding of residential rooftops.

In this work our \textbf{contributions} can be described as following:

\begin{itemize}
    \item \textit{\company Rooftop Geometry (ZRG) Dataset} - We present a novel large-scale, high quality, high resolution, multimodal dataset for residential rooftop understanding and analysis. The dataset consists of high resolution RGB orthomosaics, digital surface models (DSM), 3D rooftop geometry wireframes, and 3D colored point clouds acquired from roof inspections of over 20k residential properties across the U.S. We perform an in-depth analysis to highlight the value of our dataset.

    \item \textit{Baseline Experiments} - We provide baseline experiments for several common rooftop understanding tasks to display a few of the potential applications of our proposed dataset, including roof outline extraction, monocular height estimation, and planar roof structure extraction.
\end{itemize}

%% file: tables/datasets.tex
\begin{table*}[ht!]
\centering
\resizebox{0.8\linewidth}{!}{%
\begin{tabular}{@{}lccccccc@{}}
\toprule
\textbf{Dataset} & \textbf{Task} & \textbf{3D} & \textbf{Synthetic} & \textbf{Samples} & \textbf{Size (px)} & \textbf{Resolution (cm)} & \textbf{Modality} \\
\toprule
VWB~\cite{nauata2020vectorizing} & R & $\times$ & $\times$ & 2,001 & 256 & 30 & RGB  \\
Enschede~\cite{zhao2022extracting} & R & $\times$ & $\times$ & 2,000 & 512 & 8 & RGB \\
RID~\cite{krapf2022rid} & S & $\times$ & $\times$ & 3,648 & 512 & 10 & RGB  \\
MIPD~\cite{ren2021intuitive} & R & \checkmark & \checkmark & 2,539 & - & - & RGB,Mesh  \\
BuildingWF~\cite{luo2022learning} & R & \checkmark & \checkmark & 3,600 & - & - & RGB,Mesh  \\
\textbf{ZRG (Ours)} & S/R/H & \checkmark & $\times$ & 22,334 & 4,096+ & \textless{}1 & RGB,DSM,PC \\
\bottomrule
\end{tabular}
}
\caption{\textbf{Comparison of the ZRG dataset to related datasets} containing residential buildings and rooftop geometry related annotations. (S = segmentation, R = rooftop structure extraction, H = height estimation, PC = point cloud)}
\label{tab:datasets}
\end{table*}

%% file: sections/background.tex
While there is significant prior research into the mapping of building structure~\cite{sirko2021continental, heris2020rasterized}, building height estimation~\cite{mou2018im2height, liu2020im2elevation}, and building change~\cite{chen2020spatial, shen2021s2looking, corley2022supervising, chen2021remote} from remotely sensed imagery, the understanding of rooftop geometry and structure in particular is typically neglected from these works. Furthermore, there are few works and little focus specifically on residential buildings as opposed to commercial or industrial buildings. To further highlight the importance of automated residential rooftop understanding we detail the following notable applications:

\input{figures/roof-anatomy}

\begin{description}
    \item[Roof Damage Inspection and Detection] Automatic identification and categorization of various types of roof damage from aerial imagery~\cite{gupta2019xbd}, such as missing or damaged shingles, cracks, or leaks, enables rapid and accurate damage assessment, significantly decreasing the cost of inspections and ultimately reducing the risk of further damage to the property.

    \item[Residential Solar Rooftop Potential] Planar rooftop surface and structure extraction can assist in determining the feasibility of the installation of solar panels~\cite{krapf2022rid, suomalainen2017rooftop, assouline2017quantifying, bodis2019high, hong2017development}, by taking into account factors such as roof orientation, shading, and estimating the available surface area. Understanding the 3D geometry of roof structures allows for the optimal placement and configuration of solar panels to maximize energy production and reduce unnecessary costs to the homeowners.

    \item[3D Modeling and Digital Twins] The automatic translation of man-made structures from aerial imagery to 3D models enables the ability to simulate a realistic virtual environment for various applications such as urban planning~\cite{isikdag2009interactive} and humanitarian assistance and disaster response (HADR)~\cite{gupta2019xbd}.
\end{description}

%% file: figures/roof-anatomy.tex
\begin{figure}[ht!]
\centering
\includegraphics[width=1.0\linewidth]{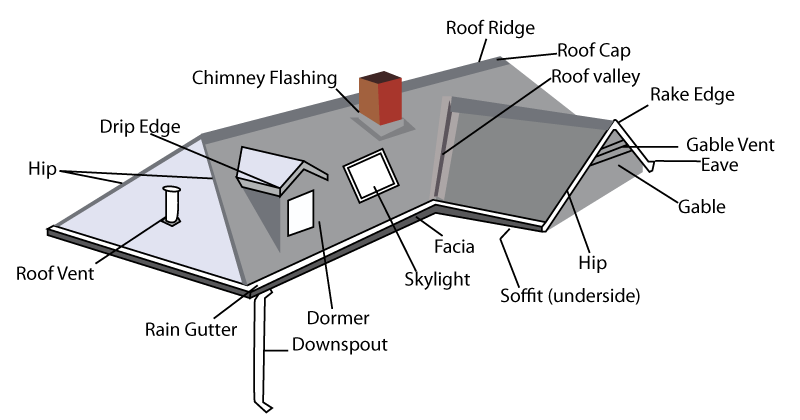}
\caption{\textbf{A visual example of the anatomy of a residential rooftop}~\cite{roof_anatomy}. Each rooftop is annotated with 3D wireframe polygons for each roof face. Individual wireframe edges are also labeled with 18 edge categories.}
\label{fig:roof-anatomy}
\end{figure}

%% file: sections/related_work.tex
\input{figures/locations}

Residential rooftop understanding datasets are scarce. The available datasets are typically generated from either multiple view reconstruction techniques~\cite{guo2022line} or from LiDAR~\cite{luo2022learning} scans which typically contain only point clouds without corresponding high resolution orthomosaics. Additionally, LiDAR comes with a high data acquisition cost, is generally low resolution, and can contain significant noise. On the other hand, aerial rooftop datasets typically do not contain any 3D modalities such as DSM or point clouds or are too low resolution to perform accurate planar rooftop structure extraction. While there are some datasets~\cite{luo2022learning,ren2021intuitive} which contain full 3D meshes of residential buildings and rooftops, they are generated using synthetic height and textures which may lead to undesirable performance in a real world roof analysis setting. The following datasets are the most closely related works in the area of residential rooftop geometry and scene understanding

\begin{description}
    \item[Vectorizing World Buildings] The VWB dataset~\cite{nauata2020vectorizing} is a modification of the SpaceNet 1 challenge dataset~\cite{van2018spacenet} which consists of 30cm spatial resolution Maxar WorldView-3 satellite RGB imagery. Patches of size 256 × 256 were manually cropped from the larger images around individual building instances. 2D planar graphs of building roof structures are annotated for 2,001 buildings from the cities of Atlanta, Paris, and Las Vegas. This dataset only contains RGB imagery and 2D wireframe annotations, but no 3D information.

    \item[Enschede] The Enschede dataset~\cite{zhao2022extracting} consists of inner and outer roofline vectors of buildings in 8cm spatial resolution aerial orthomosaics taken over the Enschede, Netherlands area. The vector annotations were extracted from the BAG dataset~\cite{bag} and overlayed onto the georeferenced imagery instead of performing manual annotation which can lead to inaccurate labels. The dataset contains 3,648 512 × 512 image patches cropped around individual buildings. This dataset only contains RGB imagery, 2D wireframe annotations, but no 3D information.

    \item[Roof Information Dataset (RID)] The RID~\cite{krapf2022rid} is a dataset of image patches centered around rooftops containing solar panel installations and is intended for photovoltaic potential analysis. This dataset contains some rooftop features within the semantic segmentation categories such as roof dormers and chimneys. However, due to the focus on solar panel detection and low spatial resolution, this dataset becomes insufficient for accurate roof structure extraction and wireframe generation. This dataset only contains RGB imagery, 2D wireframe annotations, but no 3D information.

    \item[BuildingWF] The BuildingWF dataset~\cite{luo2022learning} is composed of 3,600 polygon meshes of residential buildings with \textit{synthetic} textures along with ground truth 3D wireframes of the \textit{synthetic} meshes.

    \item[Mesh-Image Paired Dataset] The Mesh-Image paired dataset~\cite{ren2021intuitive} consists of 2,539 samples of roof geometries extracted from 2D images of residential buildings and then converted to 3D meshes using \textit{synthetic} height information.
\end{description}

\input{figures/hists}
\input{figures/results-roof-outline}

%% file: figures/locations.tex
\begin{figure}[t!]
\centering
\includegraphics[width=1.0\linewidth]{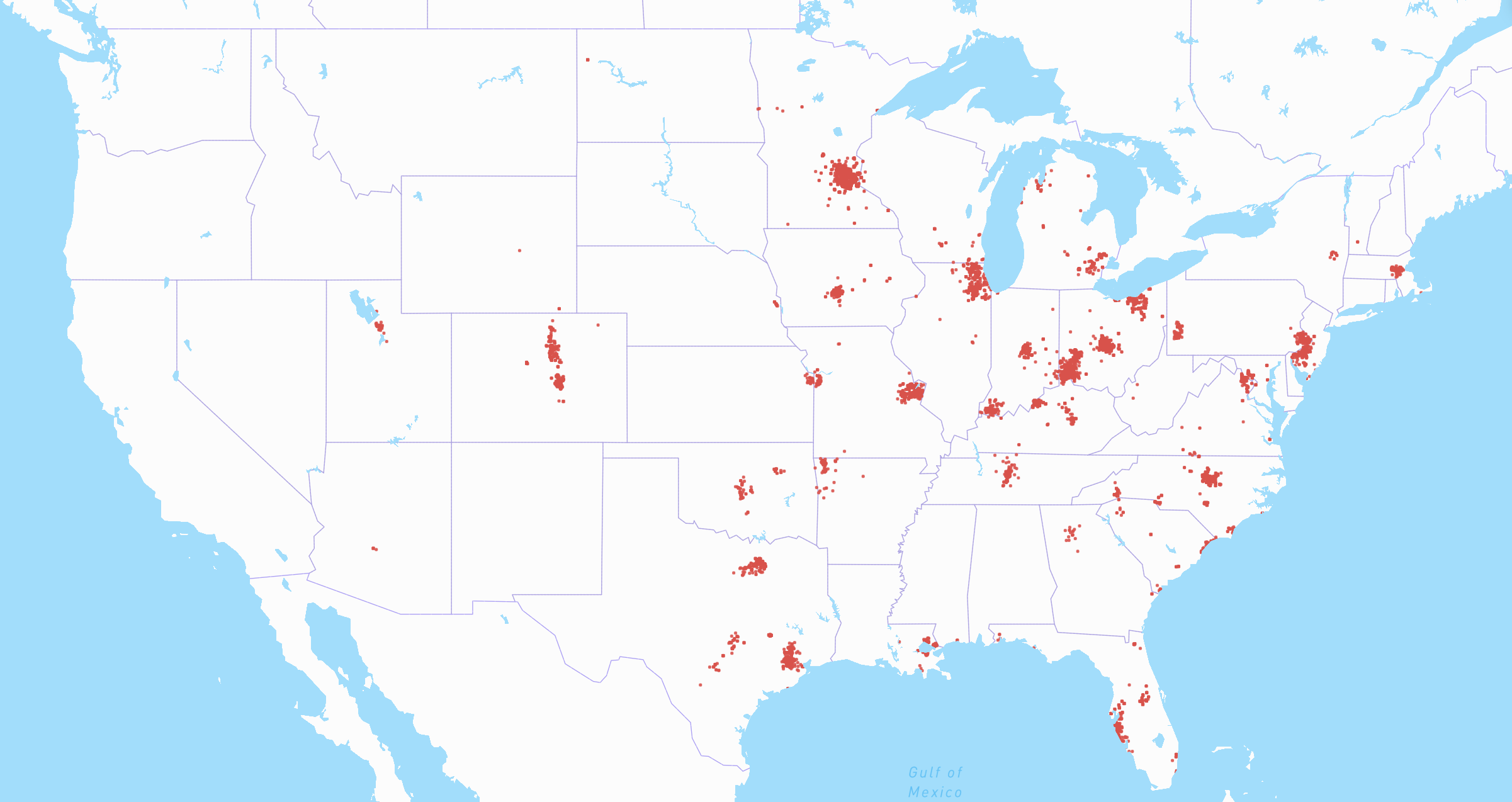}
\caption{\textbf{Residential property locations} of the ZRG dataset. The dataset consists of a diverse set of samples from various population settings (urban vs. rural), property type (single-family (SFH) vs. multi-family homes (MFH)), and regions across the U.S. (primarily northeast, south, midwest)}
\label{fig:locations}
\end{figure}

%% file: figures/hists.tex
\begin{figure}[t!]%
    \centering
    \begin{subfigure}[b]{0.49\linewidth}
        \includegraphics[width=\linewidth]{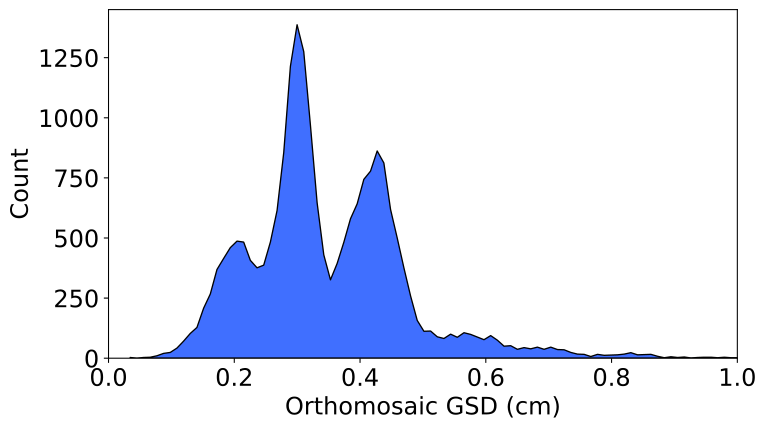}
        \caption{}
    \end{subfigure}
    \begin{subfigure}[b]{0.49\linewidth}
        \includegraphics[width=\linewidth]{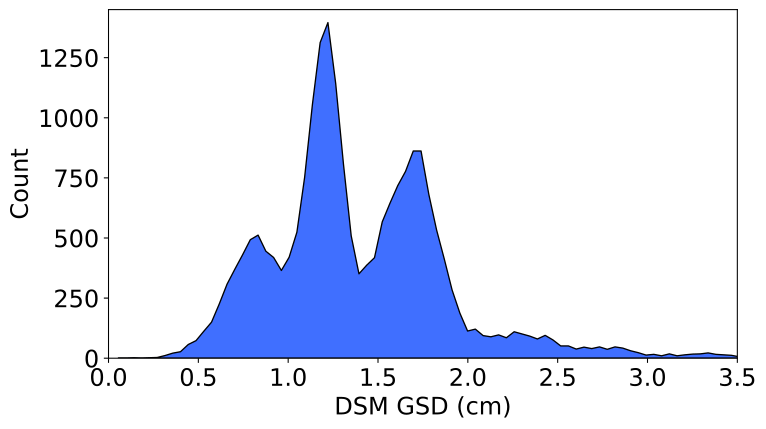}
        \caption{}
    \end{subfigure}
    \\
    \begin{subfigure}[b]{0.49\linewidth}
        \includegraphics[width=\linewidth]{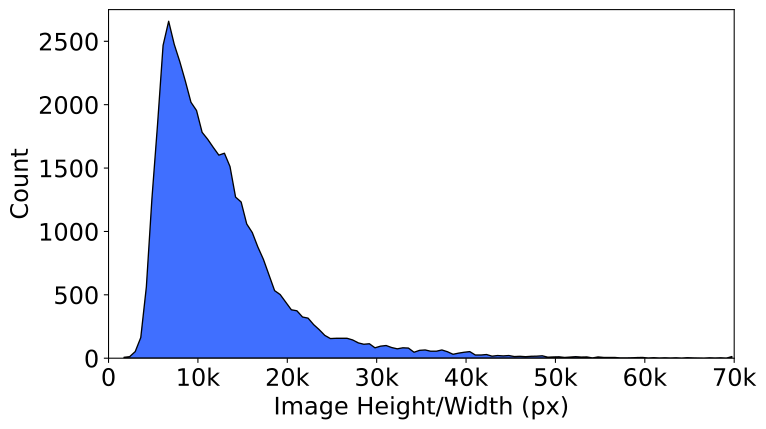}
        \caption{}
    \end{subfigure}
    \begin{subfigure}[b]{0.49\linewidth}
        \includegraphics[width=\linewidth]{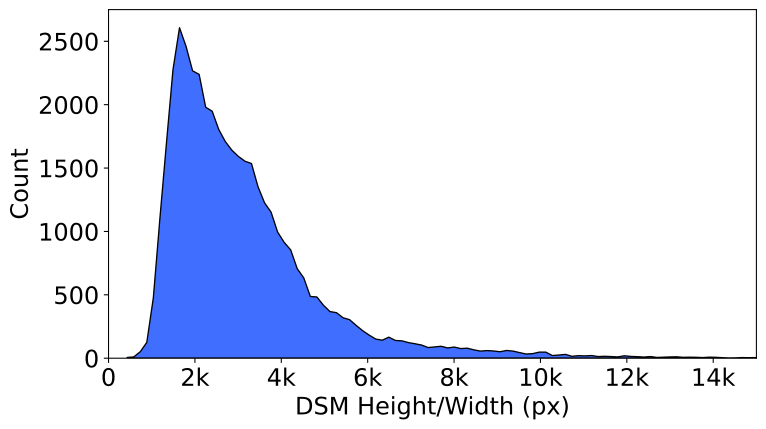}
        \caption{}
    \end{subfigure}
    \\
    \begin{subfigure}[b]{0.49\linewidth}
        \includegraphics[width=\linewidth]{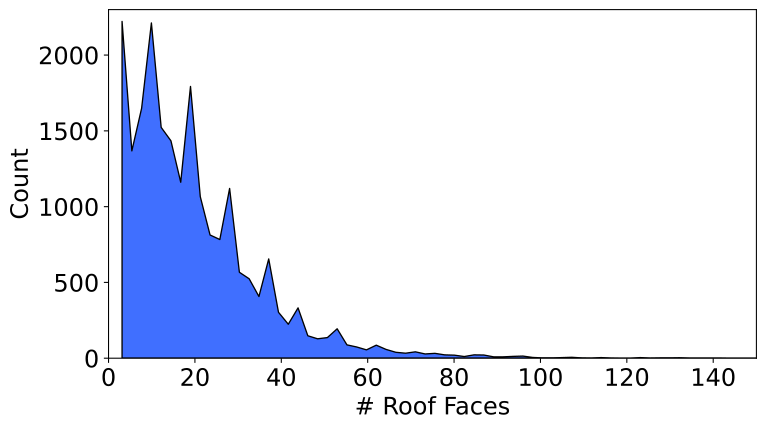}
        \caption{}
    \end{subfigure}
    \begin{subfigure}[b]{0.49\linewidth}
        \includegraphics[width=\linewidth]{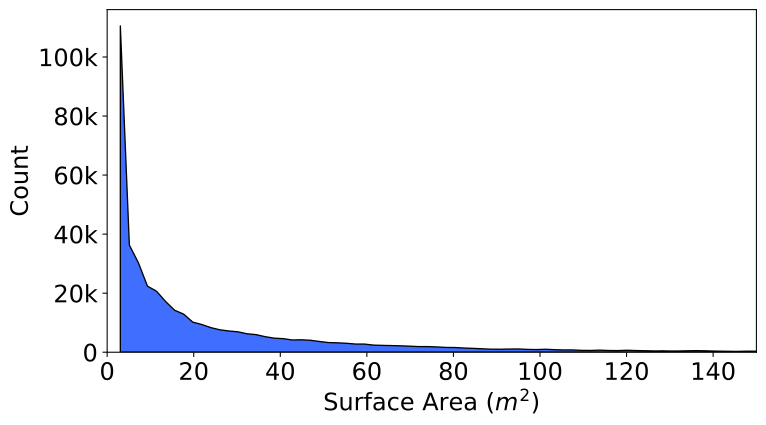}
        \caption{}
    \end{subfigure}
    \caption{Distribution plots of ground sampling distance (GSD) ($cm/px$) for (a) orthomosaic and (b) DSM, image sizes (height \& width) ($px$) for (c) orthomosaic and (d) DSM, (e) roof faces per property, and (f) surface area per roof face ($m^{2}$).}
    \label{fig:histograms}%
\end{figure}

%% file: figures/results-roof-outline.tex
\begin{figure*}[ht!]
\centering
\includegraphics[width=0.98\linewidth]{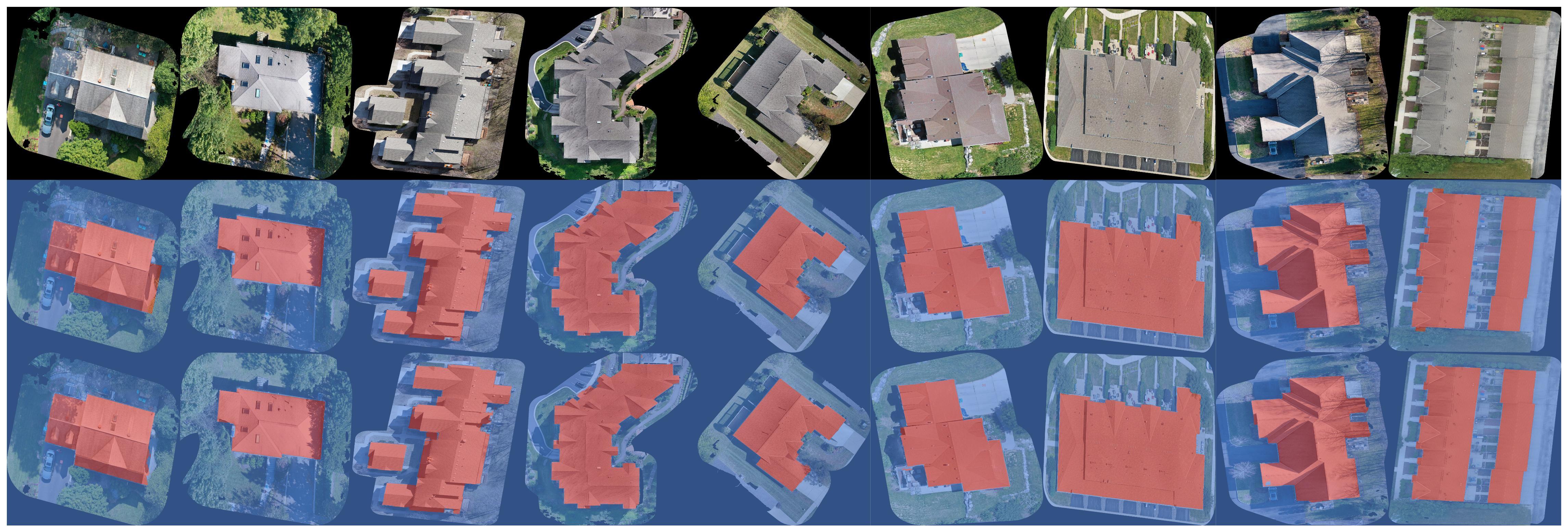}
\caption{\textbf{Roof Outline Extraction} samples and predictions from the ZRG-Test subset using the DeepLabV3-ResNet50 model trained on the ZRG-10k subset. From top to bottom: orthomosaic, ground truth, predictions. (red=\textcolor{OrangeRed}{\textbf{roof}}, blue=\textcolor{Cyan}{\textbf{background}}).}
\label{fig:results-roof-outline}
\end{figure*}

%% file: sections/dataset.tex
In this paper, we present the \company Rooftop Geometry (ZRG) dataset: a large-scale high resolution multimodal dataset with a focus on residential building rooftops for damage assessment, planar roof structure extraction, and 3D reconstruction.

\subsection{Comparison to Related Datasets}
Datasets containing residential rooftop buildings either only contain imagery but not realistic height information in the form of DSM or point clouds, or contain point clouds only but no imagery or geometric roof information. Related datasets are also generally too low resolution to support high quality and accurate understanding of roof structures. Table~\ref{tab:datasets} presents comparisons between the ZRG dataset and closely related roof structure datasets detailed in Section~\ref{sec:related-work}.

\subsection{Data Acquisition}
The data collection process utilized several commercially available DJI drones outfitted with high-resolution cameras to acquire imagery for performing residential roof inspections and analysis. The drones were programmed to navigate in a systematic lawn mower pattern, maintaining an altitude of 10-15 feet above the highest point of the roof. Additionally, oblique images were acquired for performing multi-view 3D reconstruction to infer the geometric structure of the rooftops. As illustrated in Figure~\ref{fig:locations}, data for residential properties were collected from various regions across the U.S. and include single-family homes (SFH) and multi-family homes (MFH) or apartment complexes. There is a natural concentration in clients seeking roof inspections in the central and eastern regions of the U.S. This is due to a higher concentration of hail storms occurring in these states~\cite{schaefer2004frequency}.

\subsection{Post-Processing}
Upon completion of the data acquisition phase, the captured images are stitched together and georeferenced to generate an orthomosaic. Further, Digital Surface Model (DSM) and colored point clouds were generated using 3D multi-view reconstruction techniques. Note that the techniques utilized to generate DSMs with lesser GSDs with a factor of $3-3.5$, in this case the DSM height and widths are also less than their corresponding orthomosaic. Distribution plot of the GSD and height and widths of the orthomosaics and DSMs are illustrated in Figure~\ref{fig:histograms}. The end result is a large-scale dataset of sub-centimeter resolution RGB orthomosaics, DSMs, and colored point clouds of a total of 22,334 properties.

\input{figures/edge-cases}
\input{figures/results-dsm}

\subsection{Wireframe Annotation}
Our labeling team consists of residential properties inspection domain experts. We utilize a custom annotation tool for generating 3D wireframe annotations into geojson format. Each roof face, or distinct plane on a roof, is annotated with a 3D polygon and separate line geometries for each individual edge. Additional labels such as surface area of the polygon as well as 18 distinct edge type labels are recorded in the file metadata. The 18 edge classes include common roof edge categories such as flashing, ridge, drip edge, hip, and valley. Examples of these categories are provided in Figure~\ref{fig:roof-anatomy}. A total of 425,660 total faces were annotated with a median of 16 faces per property.  Distribution plots of the roof faces per property and surface area for each roof face are provided in Figure~\ref{fig:histograms}.

\subsection{Limitations and Challenges}
\label{sec:edge-cases}
Due to the focus on the rooftop during data acquisition, there are some edge cases that arise where image acquisition stage fails to capture the entire scene surrounding the residential building. This can result in invalid pixels being present in the final stitched orthomosaic. We elect to not clean these properties from the dataset as performing inspection and analysis with models robust enough to learn even with the presence of this noise is an important and necessary task. Additionally, natural challenges such as overhanging vegetation and shadows add additional complexity for learning rooftop structure as can be seen in Figure~\ref{fig:edge-cases}.

\subsection{Dataset Subsets}
Due to the large scale of the ZRG dataset, we split the data into several subsets to make machine learning experimentation simpler and reproducible. First, 1k properties are sampled from the total dataset which we use as a holdout test set called \textbf{ZRG-Test}. Then, from the remaining ~21k we sample 10k, 1k, and 100 properties which we coin the \textbf{ZRG-10k}, \textbf{ZRG-1k}, and \textbf{ZRG-100} subsets, respectively. Since certain major cities and regions contain more samples, we perform weighted sampling by number of properties per state such that each subset will be more geospatially diverse.

\input{figures/results-face-seg}

%% file: figures/edge-cases.tex
\begin{figure}[ht!]
\centering
\includegraphics[width=0.98\linewidth]{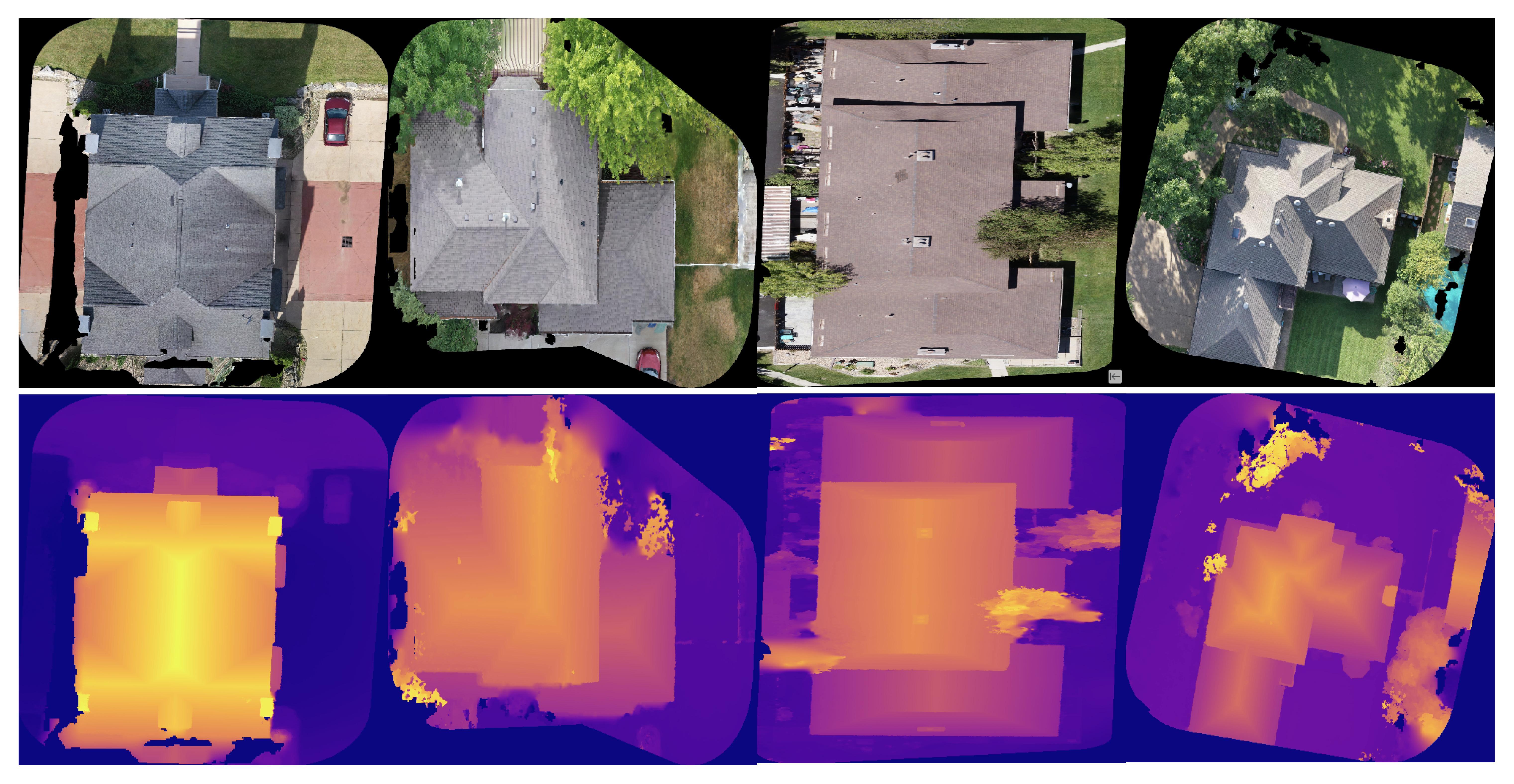}
\caption{Examples of sample images and DSM pairs containing noise as a result of invalid pixels, overhanging vegetation, and shadows.}
\label{fig:edge-cases}
\end{figure}

%% file: figures/results-dsm.tex
\begin{figure*}[ht!]
\centering
\includegraphics[width=0.98\linewidth]{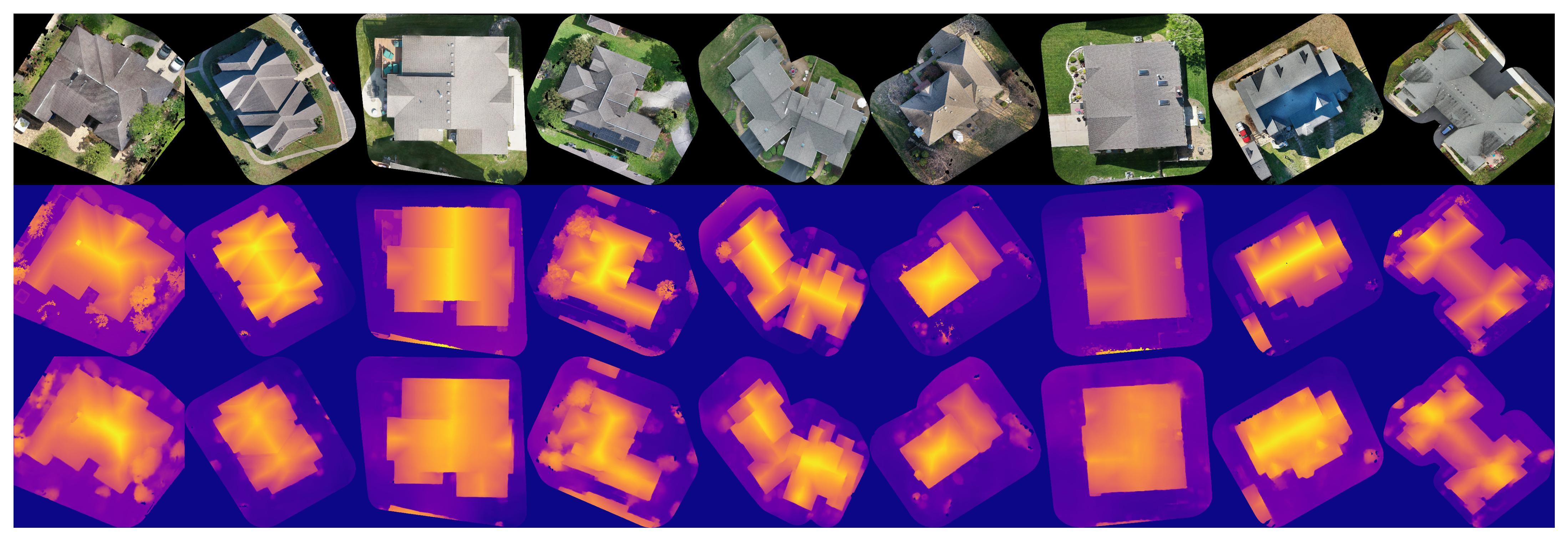}
\caption{\textbf{Monocular Height Estimation} samples and predictions from the ZRG-Test subset using the DeepLabV3-ResNet50 model trained on the ZRG-10k subset. From top to bottom: orthomosaic, ground truth DSM, height predictions. The DSM and predictions are color mapped such that \textit{brighter} indicates \textit{greater height}.}
\label{fig:results-dsm}
\end{figure*}

%% file: figures/results-face-seg.tex
\begin{figure*}[ht!]
\centering
\includegraphics[width=0.98\linewidth]{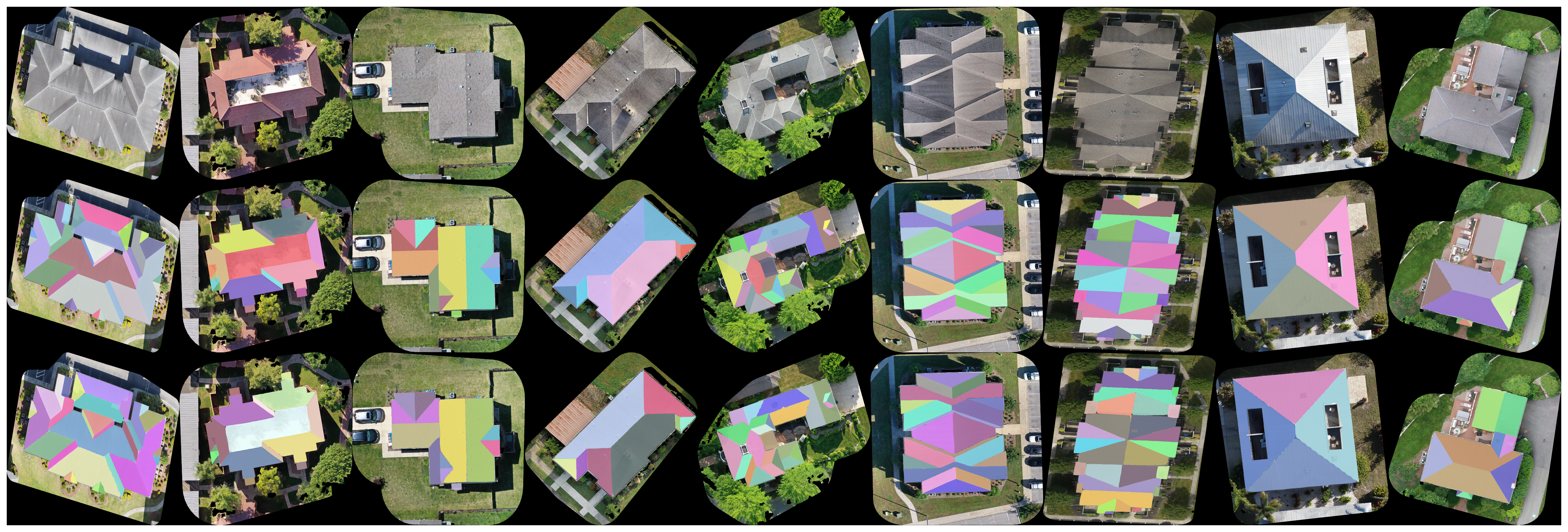}
\caption{\textbf{Planar Roof Structure Extraction} instance segmentation samples and predictions from the ZRG-Test subset using the best performing MaskRCNN model trained on the ZRG-10k subset. From top to bottom: orthomosaic, ground truth, predictions.}
\label{fig:results-face-seg}
\end{figure*}

%% file: sections/experiments.tex
We conduct several experiments to provide simple baselines using canonical architectures for the tasks of roof outline extraction, monocular height estimation, and planar roof structure extraction.

\subsection{Common Training Details}
We perform all experiments on a NVIDIA DGX server with 1x NVIDIA A100 with 40GB memory. During training we use the following augmentations: horizontal and vertical flip, random rotation, random perspective, gaussian blur, color jitter, and scale jitter~\cite{ghiasi2021simple} primarily to train models to generalize across variations in data acquisition altitude, seasonal changes, and daily changes resulting in shadows. We use the AdamW~\cite{loshchilov2017decoupled} optimizer with a learning rate of $\alpha=3e^{-4}$ throughout. We train each model using automated mixed precision (fp16) for 150 epochs with a batch size of 8 and mixed precision. Additionally, we normalize each orthomosaic using ImageNet statistics.

\subsection{Roof Outline Extraction}
\label{sec:experiments-roof-outline}
We pose the task of roof outline extraction as a binary segmentation problem where we seek to segment rooftop pixels from the background. These predictions are important for providing additional information to downstream tasks to assist in the focus on the rooftop. For this experiment we train several canonical segmentation models including U-Net~\cite{ronneberger2015u}, U-Net++~\cite{zhou2018unet++}, PSPNet~\cite{zhao2017pyramid}, and DeepLabV3+~\cite{chen2018encoder}, with ResNet~\cite{he2016deep} encoder backbones. We utilize the model implementations from the Segmentation Models PyTorch library~\cite{Iakubovskii:2019} with ImageNet pretrained weight initialization~\cite{deng2009imagenet} in the encoders. We train each model using a joint loss function consisting of the sum of the cross entropy loss and the Intersection-over-Union (IoU) loss. Evaluation is performed on the ZRG-Test holdout subset of the ZRG dataset. Random samples from the ZRG-Test set are displayed in Figure~\ref{fig:results-roof-outline} and semantic segmentation metrics for each model are provided in Table~\ref{tab:results-roof-outline}.

\subsection{Monocular Height Estimation}
\label{sec:experiments-dsm}
Estimating a surface model represented by the height of each pixel in a monocular (single) overhead view of a building is a difficult yet important problem for automated mapping purposes as explored in~\cite{mou2018im2height,xing2022sce,li2020height,liu2020im2elevation,karatsiolis2021img2ndsm,panagiotou2020generating,ghamisi2018img2dsm}. We pose this task as a dense regression problem similar to depth estimation. We utilize the same experimental setup as detailed in Section \ref{sec:experiments-roof-outline}. We min-max normalize each DSM to their relative heights using the sample statistics after filtering invalid pixels values. We train each model using a masked L1 loss function to not penalize predictions on invalid pixels. We use and modify the the segmentation model architectures described in Section \ref{sec:experiments-roof-outline} by fixing the output layer to contain a single channel with a ReLU activation~\cite{fukushima1975cognitron} for continuous outputs, similarly to the architectures used in~\cite{ranftl2021vision}. Random samples and predictions from the ZRG-Test set are displayed in Figure~\ref{fig:results-dsm} and regression metrics for each model are provided in Table~\ref{tab:results-dsm}.

\subsection{Planar Roof Structure Extraction}
\label{sec:experiments-face-seg}
For the task of planar roof structure extraction, we experiment with instance segmentation architectures for segmenting each individual face of the roof from a single overhead view. We experiment by fine-tuning the torchvision~\cite{torchvision2016} implementation of the MaskRCNN architecture~\cite{he2017mask} which is pretrained on the MS-COCO dataset~\cite{lin2014microsoft}. Random samples from the ZRG-Test set are displayed in Figure~\ref{fig:results-face-seg}. In Table~\ref{tab:results-face-seg} We report mean Average Precision (mAP) at different IoU thresholds and at different scales, (medium and large roof faces based on area). For simplicity, we pose the single-view planar roof structure extraction task as an instance segmentation problem. However, we note that it is also common to utilize corner and junction detection methods~\cite{zhao2022extracting} in combination with graph neural network (GNN) based architectures~\cite{bronstein2017geometric} as an alternative to solve this problem.

\input{tables/results-face-seg}

%% file: tables/results-face-seg.tex
\begin{table}[ht!]
\centering
\resizebox{0.98\linewidth}{!}{%
\begin{tabular}{@{}lccccc@{}}
\toprule
\textbf{Subset} &
\multicolumn{1}{c}{\textbf{mAP}} &
\multicolumn{1}{c}{\textbf{$\text{mAP}_{50}$}} &
\multicolumn{1}{c}{\textbf{$\text{mAP}_{75}$}} &
\multicolumn{1}{c}{\textbf{$\text{mAP}_{M}$}} &
\multicolumn{1}{c}{\textbf{$\text{mAP}_{L}$}} \\
\toprule
ZRG-100 & 40.4 & 45.3 & 35.5 & 37.1 & 66.3 \\
ZRG-1k & 67.9 & 91.4 & 44.5 & 66.4 & 92.4 \\
ZRG-10k & \textbf{72.1} & \textbf{97.0} & \textbf{47.1} & \textbf{90.0} & \textbf{96.1} \\
\bottomrule
\end{tabular}%
}
\caption{
\textbf{Planar Roof Structure Extraction} instance segmentation mean Average Precision (mAP) results of a MaskRCNN with a ResNet50-FPN backbone pretrained on COCO and trained on each ZRG subset and evaluated on the ZRG-Test subset. Best results marked in \textbf{bold}.}
\label{tab:results-face-seg}
\end{table}

%% file: sections/discussion.tex
\input{tables/results-dsm}

\subsection{Effects of Dataset Size}
\label{sec:experiments-data-size}
To explore the necessity of creating a large scale dataset, we repeat the experiments in Section~\ref{sec:experiments-face-seg} for planar roof structure extraction on the ZRG-100, ZRG-1k, and ZRG-10k subsets to evaluate how dataset size affects performance. We train each method for 1k iterations and record the mean Average Precision (mAP) at different thresholds and for different size roof faces (medium (M) and large (L)). As seen in Table~\ref{tab:results-face-seg}, the increasing size of the dataset results in a significant increase in performance across all metrics with a particularly large increase in $AP_{M}$ performance in for medium sized roof faces. This illustrates the need for large-scale and high quality labeled datasets to provide greater performance gains rather than iterating with various model architectures.

\subsection{Results}
With regards to the roof outline extraction and monocular view height estimation tasks, it is clear that the DeepLabV3+ model with a ResNet50 backbone outperforms other models in all cases. Predictions in Figures~\ref{fig:results-roof-outline} and \ref{fig:results-dsm} visually show that the model is able to properly segment the roof from the background and accurately estimate pixelwise height of the buildings.

For planar roof structure extraction, we can see that there is an increasing relationship between the performance and the size of the training set, particularly with a large increase of $23.6$ in mAP for medium sized roof faces when increasing the dataset size from 1k to 10k samples. Additionally, Figure~\ref{fig:results-face-seg} shows that the model is able to extrapolate the segmentation of roof face structures even in the presence of occlusion of areas of the face due to overhanging vegetation.

\input{tables/results-roof-outline}

\subsection{Future Work}
While each task may not appear to be as useful independently, the bigger picture of combining predicted roof outlines, height estimations, and segmented roof faces to generated 3D reconstructions and wireframes of rooftops allows for deeper analysis and insights into the condition and surface area breakdown of each roof. However, we leave this combination of model outputs or joint learning for future work.

We plan to perform further annotation of the dataset to include labels for classification of roof types~\cite{mohajeri2018city,buyukdemircioglu2021deep,alidoost2018cnn}, e.g. gable, complex, pyramidal, as well as labels for classification of building type, particularly single vs. multi-family homes, similar to the work described in ~\cite{bandam2022classification}. Both label categories can be used to further benefit fine-grained analysis of rooftops.

While we do not explicitly utilize the multi-view imagery used to generate the DSM and point cloud, we acknowledge that there are numerous recent reconstruction methods related to Neural Radiance Fields (NeRF)~\cite{mildenhall2021nerf,zhang2020nerf++,yu2021pixelnerf} which can more accurately generate 3D models and novel views of each property.

Due to the regional diversity of the properties, our dataset can result in subpopulation distribution shifts described in~\cite{koh2021wilds,sagawa2021extending}. Further analysis of the generalization and geographic bias across regions of the U.S. and distance from urban metropolitan areas is outside the scope of this paper and we leave for future work.

%% file: tables/results-dsm.tex
\begin{table}[ht!]
\centering
\resizebox{1.0\linewidth}{!}{%
\begin{tabular}{ccccc}
\toprule
\textbf{Model} &
\textbf{Backbone} &
\textbf{MAE} &
\textbf{MSE} &
\textbf{RMSE} \\
\toprule
PSPNet & ResNet18 & 0.0815 & 0.0167 & 0.1292 \\
U-Net & ResNet18 & 0.0738 & 0.0144 & 0.1198 \\
U-Net++ & ResNet18 & 0.0727 & 0.0138 & 0.1174 \\
DeepLabV3+ & ResNet18 & \textbf{0.0696} & \textbf{0.0131} & \textbf{0.1143} \\
\midrule
PSPNet & ResNet50 & 0.0793 & 0.0162 & 0.1274 \\
U-Net & ResNet50 & 0.0723 & 0.0133 & 0.1155 \\
U-Net++ & ResNet50 & 0.0699 & 0.0127 & 0.1129 \\
DeepLabV3+ & ResNet50 & \textbf{0.0679} & \textbf{0.0123} & \textbf{0.1107} \\ 
\bottomrule
\end{tabular}%
}
\caption{
\textbf{Monocular Height Estimation results} of various models trained on the ZRG-10k subset and evaluated on the ZRG-Test subset. We report mean absolute error (MAE), mean squared error (MSE), and root mean squared error (RMSE) dense regression metrics. Metrics are computed on the relative height values after min-max normalization. Best results are marked in \textbf{bold}.}
\label{tab:results-dsm}
\end{table}

%% file: tables/results-roof-outline.tex
\begin{table}[ht!]
\centering
\resizebox{0.95\linewidth}{!}{%
\begin{tabular}{ccccc}
\toprule
\textbf{Model} &
\textbf{Backbone} &
\textbf{OA}  &
\textbf{F1} &
\textbf{mIoU} \\
\toprule
PSPNet & ResNet18 & 92.37 & 90.37 & 76.99 \\
U-Net & ResNet18 & 96.89 & 96.28 & 92.93 \\
U-Net++ & ResNet18 & 97.59 & 97.15 & 95.91 \\
DeepLabV3+ & ResNet18 & \textbf{97.73} & \textbf{97.32} & \textbf{96.65} \\
\midrule
PSPNet & ResNet50 & 96.48 & 95.80 & 92.52 \\
U-Net & ResNet50 & 97.52 & 97.06 & 95.78 \\
U-Net++ & ResNet50 & 97.31 & 96.82 & 95.12 \\
DeepLabV3+ & ResNet50 & \textbf{97.81} & \textbf{97.42} & \textbf{96.83} \\ 
\bottomrule
\end{tabular}%
}
\caption{
\textbf{Roof Outline Extraction results} of models trained on the ZRG-10k subset and evaluated on the ZRG-Test subset. We report overall accuracy (OA), average F1 score, and mean Intersection-over-Union (mIoU) semantic segmentation metrics. Best results are marked in \textbf{bold}.}
\label{tab:results-roof-outline}
\end{table}

%% file: sections/conclusion.tex
In this paper, we presented ZRG, a 3D residential rooftop understanding dataset, which we have shown through thorough analysis and several baseline experiments what is possible with a large-scale dataset with multiple modalities. We hope that our work advances and generates novel ideas for additional applications of residential rooftop structure extraction and understanding and inspires the research community to develop additional residential rooftop datasets.